\documentclass[10pt,twocolumn,letterpaper]{article}

\usepackage{iccv}
\usepackage{times}
\usepackage{epsfig}
\usepackage{graphicx}
\usepackage{amsmath}
\usepackage{amssymb}
\usepackage{bbm}
\usepackage{enumitem}
\usepackage{tabularx}
\usepackage{rotating}
\usepackage{xcolor}
\usepackage{subcaption}
\usepackage{multirow}
\usepackage{booktabs}
\usepackage{url}
\usepackage[font=small]{caption}

\newcommand*\rot{\rotatebox{90}}
\newcommand\boldhead[1]{\vspace{0.03in}\noindent\textbf{#1: }}
\iccvfinalcopy % *** Uncomment this line for the final submission

\def\iccvPaperID{1997} % *** Enter the ICCV Paper ID here

\ificcvfinal\fi
\begin{document}

%%%%%%%%% TITLE
\title{Aligned Image-Word Representations Improve Inductive Transfer Across Vision-Language Tasks}
%\title{Improving Inductive Transfer across Vision-Language Tasks through Aligned Image-Word Representations}
%\title{Learning Image-Word Representations from Complementary and Hierarchical Vision-Language Tasks}
% Keywords - Shared/rich/generalizable representation, multiple vision-language tasks, inductive transfer, intermediate supervision 
% Sharing Vision-Language representations across multiple tasks for inductive transfer   
% What does sharing Vision-Language representation across tasks give us?
% Learning image and language representations from diverse and complementary experience
% Learning image and language representations from diverse and complementary task hierarchy
%\and
\author{Tanmay Gupta$^1$ \hspace{0,5cm} Kevin Shih$^1$ \hspace{0,5cm} Saurabh Singh$^2$ \hspace{0,5cm} Derek Hoiem$^1$\\
$^1$University of Illinois, Urbana-Champaign  \hspace{0,5cm} $^2$Google Inc.\\
{\tt\small \{tgupta6, kjshih2, dhoiem\}@illinois.edu} \hspace{0,5cm} {\tt\small saurabhsingh@google.com}}

\maketitle
%\thispagestyle{empty}

%%%%%%%%% ABSTRACT
\begin{abstract}
   An important goal of computer vision is to build systems that learn visual representations over time that can be applied to many tasks. In this paper, we investigate a vision-language embedding as a core representation and show that it leads to better cross-task transfer than standard multi-task learning. In particular, the task of visual recognition is aligned to the task of visual question answering by forcing each to use the same word-region embeddings. We show this leads to greater inductive transfer from recognition to VQA than standard multitask learning. Visual recognition also improves, especially for categories that have relatively few recognition training labels but appear often in the VQA setting.  Thus, our paper takes a small step towards creating more general vision systems by showing the benefit of interpretable, flexible, and trainable core representations.

\end{abstract}

%%%%%%%%% BODY TEXT
\vspace{-5mm}

\section{Introduction}

Consider designing a vision system that solves many tasks. Ideally, any such system should be able to reuse representations for different applications. As the system is trained to solve more problems, its core representations should become more complete and accurate, facilitating the learning of additional tasks. Vision research often focuses on designing good representations for a given task, but what are good core representations to facilitate learning the next?

The application of knowledge learned while solving one task to solve another task
%The phenomenon of reusing knowledge learned while solving one task and applying it to another task
is known as \textit{transfer learning} or \textit{inductive transfer}. Inductive transfer has been demonstrated in recent vision-language tasks in~\cite{li2016learning,lu2016hierarchical,jabri2016revisiting,fukui2016multimodal,shih2016look,ilievski2016focused}, where the hidden or output layers of deep networks learned from pre-training (e.g. on ImageNet~\cite{deng2009imagenet}) or multitask learning serve as the foundation for learning new tasks. However, the relations of features to each new task needs to be re-learned using the new task's data. The goal of our work is to transfer knowledge between related tasks \emph{without} the need to re-learn this mapping. Further, as we are working with vision-language tasks, we aim to transfer knowledge of both vision \emph{and} language across tasks.

\begin{figure}[t]

\centering
\includegraphics[width=\linewidth]{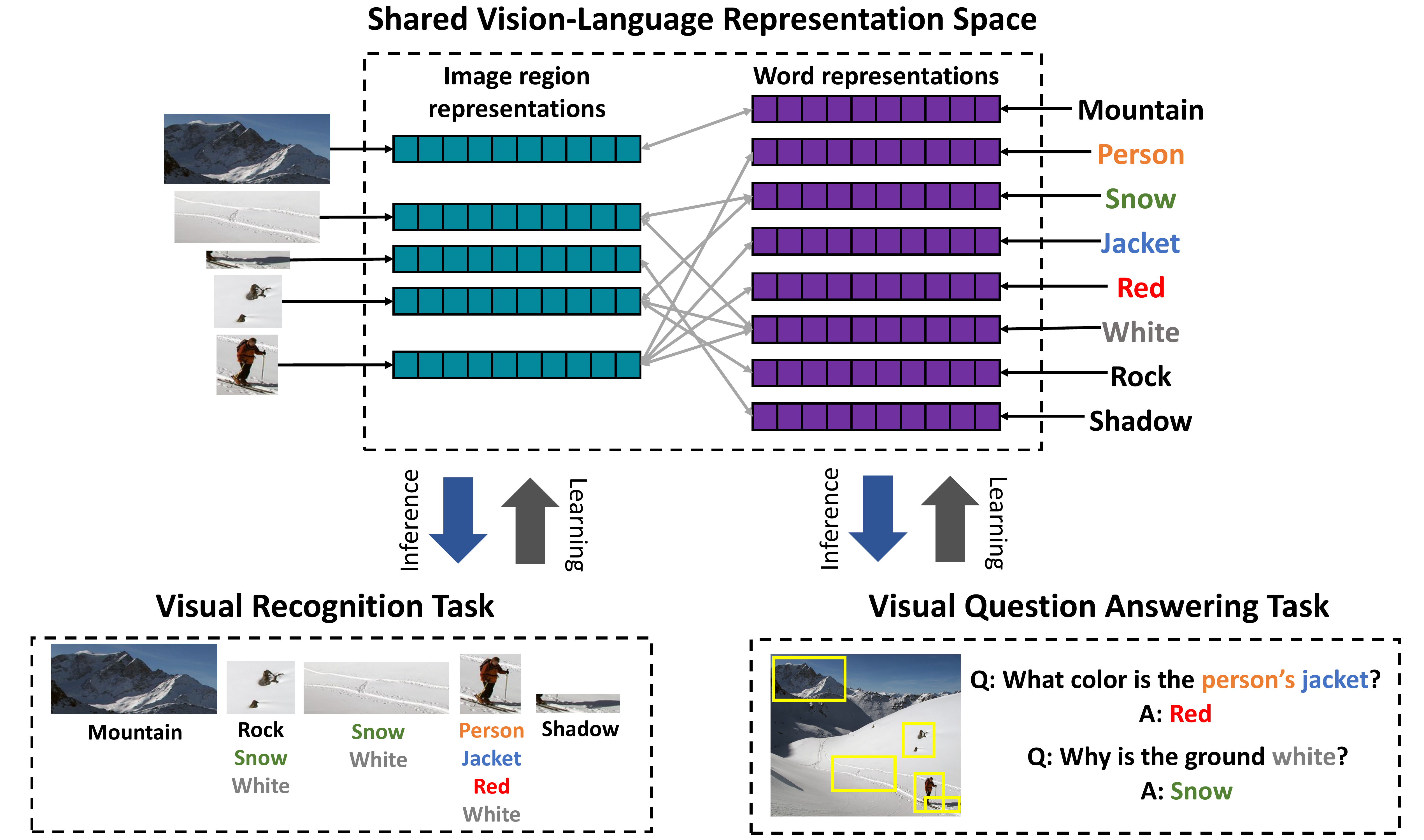}
\vspace{-7mm}
\caption{\textbf{Sharing image-region and word representations across multiple vision-language domains:} The SVLR module projects images and words into a shared representation space. The resulting visual and textual embeddings are then used for tasks like Visual Recognition and VQA. The models for individual tasks are formulated in terms of inner products of region and word representations enforcing an alignment between them in the shared space.}
\vspace{-6mm}
\label{fig:short_overview}
\end{figure}

% give high level approach now, talk about parsing details mapping question to noun and adj in the methods section
In this work we propose a Shared Vision-Language Representation (SVLR) module that improves inductive transfer between related vision-language tasks (see Fig.~\ref{fig:short_overview}).  We apply our approach to visual recognition (VR) and attention-based visual question answering (VQA). We formulate VR in terms of
a joint embedding of textual and visual representations computed by the SVLR module. Each region is mapped closest to its correct (textual) class label.  For example, the embedding of ``dog'' should be closer to an embedded region showing a dog than any other object label.  We formulate VQA as predicting an answer from a relevant region, where relevance and answer scores are computed from embedded word-region similarities.  For example, a region will be considered relevant to ``Is the elephant wearing a pink blanket?'' if the embedded ``pink'' \textit{and} either ``elephant'' or ``blanket'' are close to the embedded region. Similarly, the answer score considers embedded similarities, but in a more comprehensive manner. We emphasize that the same word-region embedding is learned for both VR and VQA.  Our experiments show that formulating both tasks in terms of the SVLR module leads to better cross-task transfer than if features are shared through multitask learning but without exploiting the alignment between words and regions. 

In summary, \textbf{our main contribution} is to show that the proposed SVLR module leads to better inductive transfer than unaligned feature sharing through multitask learning. As an added benefit, \textit{attention} in our VQA model is highly interpretable: we can show what words cause the system to score a particular region as relevant. We take a small step towards lifelong-learning vision systems by showing the benefit of an interpretable, flexible, and trainable core representation.  

\begin{figure*}[t]
\begin{center}
%\fbox{\rule{0pt}{2in} \rule{.6\linewidth}{0pt}}
\includegraphics[width=\linewidth]{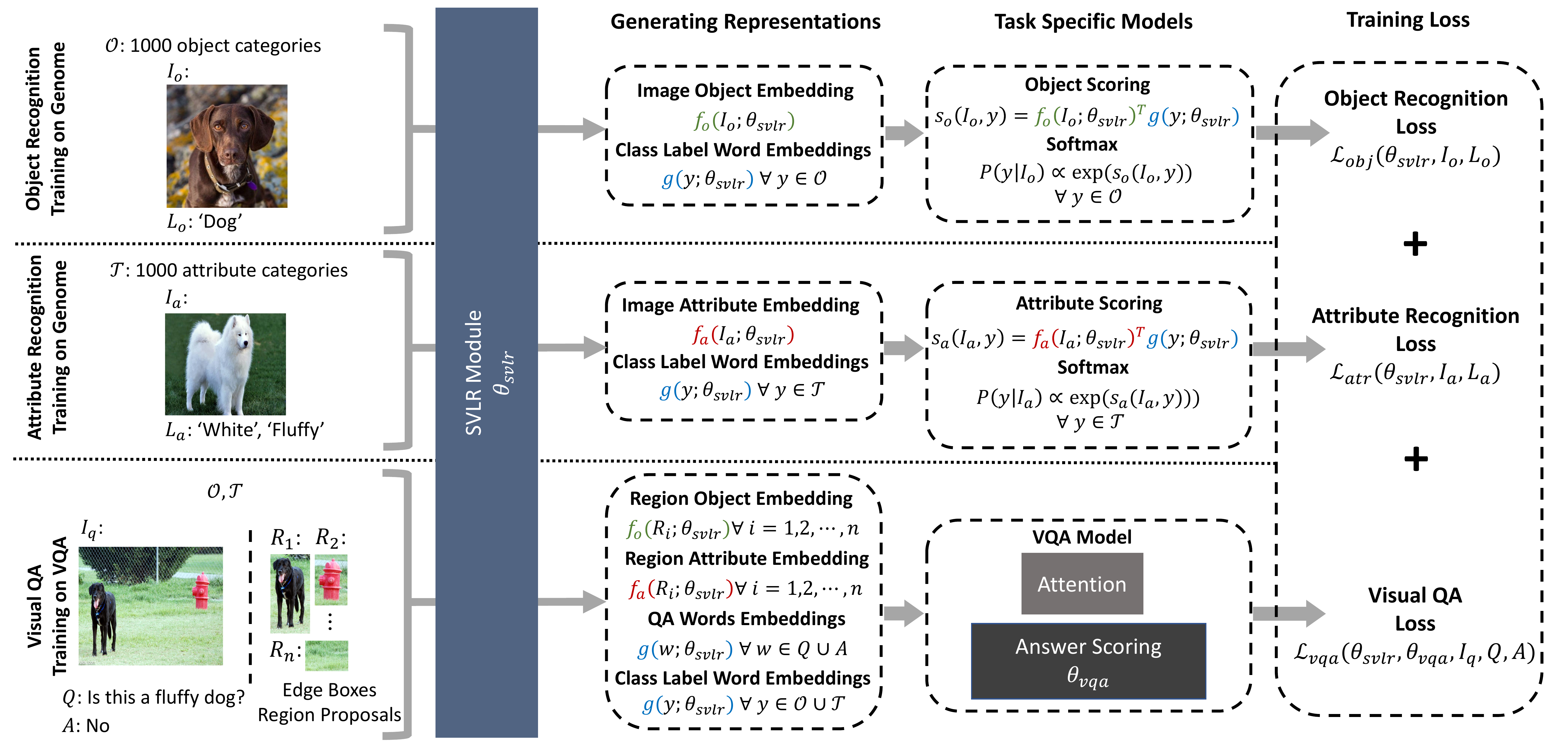}
\vspace{-0.9cm}
\end{center}
   \caption{\textbf{Joint Training on Visual Recognition(VR) and Visual Question Answering(VQA) with the proposed SVLR Module:} The figure depicts sharing of image and word representations through the SVLR module during joint training on object recognition, attribute recognition, and VQA. The recognition tasks use object and attribute labelled regions from Visual Genome while VQA uses images annotated with questions and answers from the VQA dataset. The benefit of joint training is that while the VQA dataset does not provide region groundings of nouns and adjectives in the QA (e.g. ``fluffy",``dog"), this complementary supervision is provided by the Genome recognition dataset. Models for each task involve image and word embeddings produced by SVLR module or their inner products (See Fig~\ref{fig:system} for VQA model architecture).
   }
   \vspace{-6mm}
\label{fig:features}
\end{figure*}

%  \caption{\textbf{Joint Training on Visual Recognition(VR) and Visual Question Answering(VQA) with the proposed SVLR Module:} The figure depicts sharing of image and word representations through the SVLR module during joint training on object recognition, attribute recognition, and VQA. The recognition tasks use object and attribute labelled regions from Visual Genome while VQA uses images annotated with questions and answers from the VQA dataset. The benefit of joint training is that while the VQA dataset does not provide region groundings of nouns and adjectives in the QA (eg. 'fluffy','dog'), this complementary supervision is provided by the Genome recognition dataset. Models for each task involve image and word embeddings produced by SVLR module or their inner products (See Fig~\ref{fig:system} for VQA model architecture). The parameters of the SVLR module and Answer Scoring module in VQA are trained using stochastic gradient descent. As a baseline, we replace $g(y;\theta_{svlr})$, a function of word2vec representation of $y$, with learnable weight vector $h_y$ in the VR models. This is equivalent to using a 1000 way classification layer like the last layers of AlexNet, VGG, or ResNet instead of using our language embeddings for classification. This enables comparison of inductive transfer due to joint training both with and without sharing language representations.}

%-------------------------------------------------------------------------
\section{Related Work}
\label{sec:related}
\noindent \textbf{Never-ending learning:}  NEL~\cite{mitchell2010never,carlson2010toward,thrun1998lifelong,silver2013lifelong,chen2013iccv} aims to continuously learn from multiple tasks such that learning to solve newer problems becomes easier. Representation learning~\cite{bengio2013pami}, multitask learning~\cite{caruana1998springer}, and curriculum learning~\cite{pentina2015curriculum} are different aspects of this larger paradigm. Inductive transfer through shared representations is a necessary first step for NEL. Most works focus on building transferable representations within a single modality such as language or vision only. We extend this framework to learn a joint vision-language representation which enables a much larger class of new vision-language tasks to easily build on and contribute to the shared representation. \\
\vspace{-2mm}

\noindent \textbf{VR using Vision-language embeddings:} Traditionally, visual recognition has been posed as multiclass classification over discrete labels~\cite{he2015deep,simonyan2014very,krizhevsky2012imagenet}. Using these recognizers for tasks like VQA and image captioning is challenging because of the open-vocabulary nature of these problems. However, availability of continuous word embeddings (e.g. word2vec~\cite{mikolov2013efficient}) has allowed reformulation of visual recognition as a nearest neighbor search in a learned image-language embedding space~\cite{wang2016learning}. Such embeddings have been successfully applied to a variety of tasks that require recognition such as image captioning~\cite{lin2014microsoft,hodosh2013framing}, phrase localization~\cite{plummer2015flickr30k,krishna2016visual}, referring 
expressions~\cite{KazemzadehOrdonezMattenBergEMNLP14,Mao2016cvpr}, and VQA \cite{antol2015vqa,ren2015nips,yu2015visual}. 

Our recognition model is related to previous open-vocabulary recognition/localization models ~\cite{wang2016learning,rohrbach2016grounding,gong2014improving}, which learn to map visual CNN features to continuous word vector representations. However, we specifically focus on the multitask setting where VR forms a part of a higher-level vision-language task such as VQA. Since the SVLR module is reused in both tasks with inner products in the embedding space forming the basis for both models, during joint training VQA provides a weak supervision for recognition as well. Fang~\etal~\cite{fang2015cvpr} also learn object and attribute classifiers from weak supervision in the form of image-caption pairs using a multiple instance learning (MIL) framework, but do not use a vision-language embedding. Liu~\emph{et al.}~\cite{liu2016attention} similarly use VR annotation from Flickr30K entities~\cite{plummer2015flickr30k} to co-supervise attention in a caption-generation model on the same dataset. Our work goes further by allowing the supervision to come from separate datasets, thereby increasing the amount of training data available for the shared parameters. Additionally, we look at how each task has benefited from jointly training with the other.\\
\vspace{-2mm}

%{While the classifier weights can be interpreted as language embeddings, they are not reused as word representations during caption generation. We note that it is trivial to incorporate their noisy-OR MIL loss in our framework and could yield performance improvements.}

%{But there are two main differences. First, instead of trying to create language embeddings that summarily accommodate words and phrases (eg. by using the hidden state of an RNN), we learn single-word representations and construct more complex phrase representations by averaging constituent word representations (sentence representation is described in \todo{method section}). Second, since object and attribute class labels in the VR task are also part of QA annotations for VQA, we use the same word representations for VQA allowing them to receive additional training signals from the VQA task.}

\noindent \textbf{VQA:} Visual Question Answering (VQA) involves responding to a natural language query about an image. Our VQA model is closely related to attention-based VQA models~\cite{fukui2016multimodal,ilievski2016focused,lu2016hierarchical,xu2016ask,shih2016look,yang2015stacked,andreas2016neural,andreas2016learning, kumar2015ask,tommasi2016bmvc} which attempt to compute a distribution (region relevance or \textit{attention}) over the regions/pixels in an image using inner product of image-region and the \textit{full} query embedding~\cite{xu2016ask, shih2016look, ilievski2016focused,lu2016hierarchical}. Region relevance is used as a weight to pool relevant visual information which is usually combined with the language representation to create a multimodal representation. 
%Finally, this representation is passed through a classifier producing an answer score.
Various methods of pooling such as elementwise-addition, multiplication, and outer-products have been explored~\cite{yang2015stacked, fukui2016multimodal}. Attention models are themselves an active area of research with applications in visual recognition~\cite{mnih2014nips,jaderberg2015nips}, object localization, caption generation~\cite{johnson2015arxiv}, question answering~\cite{weston2014memory,sukhbaatar2015end,kumar2015ask}, machine comprehension~\cite{hermann2015teaching} and translation ~\cite{bahdanau2014arxiv,wu2016arxiv}, and neural turing machines ~\cite{graves2014arxiv}.
  
Our model explicitly formulates attention in VQA as image localization of nouns and adjectives mentioned in a candidate QA pair. Ilievski~\etal~\cite{ilievski2016focused} use a related approach for attention. They use word2vec to map individual words in the question to the class labels of a pre-trained object detector which then generates the attention map by identifying regions for those labels. Tommasi~\etal~\cite{tommasi2016bmvc} similarly use a pre-trainined CCA \cite{gong2014improving} vision-language embedding model to localize noun phrases, then extracts scene, attribute, and object features to answer VQA questions. Our model differs from these methods in two ways: (i) vision-language embeddings for VR allow for end-to-end trainability, and (ii) jointly training on VR provides additional supervision of \textit{attention} through a different (non-VQA) dataset.

%Another work that relies heavily on the question parse is Andreas \etal~\cite{andreas2016neural,andreas2016learning} which uses the syntactic parse to dynamically arrange a set of parametrized neural modules to form a larger question-specific VQA model. Each module performs a specific function such as localizing a specific word or verifying relative locations such as above and below. In contrast, our model is static and simpler while making use of language parsing to make the model interpretable and modular.  

Andreas~\etal~\cite{andreas2016neural,andreas2016learning} rely heavily
on the syntactic parse to dynamically arrange a set of parametrized neural modules. Each module performs a specific function such as localizing a specific word or verifying relative locations. In contrast, our approach uses a static model but relies on language parse to make it interpretable and modular.

%-------------------------------------------------------------------------

\section{Method}
We propose an SVLR module to facilitate greater inductive transfer across vision-language tasks. As shown in Fig.~\ref{fig:features}, the word and region representations required for object recognition, attribute recognition, and VQA are computed through the SVLR module. By specifically formulating each task in terms of inner products of word and region representations and training on all tasks jointly, we ensure each task provides a consistent, non-conflicting training signal for aligning words and region representations. During training, the joint-task model is fed batches containing training examples from each task's dataset.

\vspace{-1mm}

\subsection{Shared Vision Language Representation}
\label{sec:embedding}
The SVLR module converts words and image-regions into feature representations that are aligned to each other and \textit{shared} across tasks.  \\

\noindent\textbf{Word Representations:} The representation $g(w)$ for a word $w$ is constructed by applying two fully connected layers (with 300 output units each) to pretrained word2vec representation~\cite{word2vec} of $w$ with ReLU after the first layer. \\

\noindent\textbf{Region Representations:} A region $R$ is represented using two $300$ dimensional feature vectors $f_o(R)$ and $f_a(R)$ that separately encode the objects and attributes contained. We used two representations instead of one to encourage disentangling of these two factors of variation. For example, we do not expect ``red" to be similar to ``apple", but we expect $f_o(R)$ and $f_a(R)$ to be similar to $g(``red")$ and $g(``apple")$ if $R$ depicts a red apple. The features are constructed by extracting the average pooled features from Resnet~\cite{he2015deep} pretrained on ImageNet and then passing through separate object and attribute networks. Both networks consist of two fully connected layers (with 2048 and 300 output units) with batch normalization~\cite{batchnorm} and ReLU activations. 

\subsection{Visual Recognition using SVLR}
\label{sec:vr}
\subsubsection{Inference}\label{sec:recog_inference}
The visual recognition task is to classify image regions into one or more object and attribute categories. The classification score for region $R$ and object category $w$ is $f_o^T(R)g(w)$. The classification score for an attribute category $v$ is $f_a^T(R)g(v)$. Attributes may include adjectives and adverbs (e.g., ``standing''). Though our recognition dataset has a limited set of object categories $\mathcal{O}$ and attribute categories $\mathcal{T}$,  our model can produce classification scores for any object or attribute label given its word2vec representation. In experiments, the $\mathcal{O}$ and $\mathcal{T}$ consist of 1000 most frequent object and attribute categories in the Visual Genome dataset~\cite{krishna2016visual}. 

\vspace{-3mm}
\subsubsection{Training}\label{sec:recog_learn}
Our VR model is trained using the Visual Genome dataset which provides image regions annotated with object and attribute labels. VR uses only the parameters for the embedding functions $f_o, f_a$ and $g$ that are part of the SVLR module. The parameters of $f_o$ receive gradients from the object loss while those of $f_a$ receive gradients from the attribute loss. The parameters of word embedding model $g$ receive gradients from both losses.\\

\noindent 
\textbf{Object loss:} %We use a multi-label classification loss as object classes may not be mutually exclusive due to hypernyms (e.g., ``man'' \emph{is a} ``person'') and synonyms.  
We use a multi-label loss as object classes may not be mutually exclusive (e.g., ``man'' \emph{is a} ``person''). For a region $R_j$, we denote the set of annotated object categories and their hypernyms extracted from WordNet \cite{miller1995acm} by $\mathcal{H}_j$. The object loss forces the true labels and their hypernyms to score higher than all other object labels by a margin $\eta_{obj}$. For a batch of $M$ samples $\{(R_j,\mathcal{H}j)\}_{j=1}^{M}$ the object loss is:
\vspace{-5mm}
\begin{multline}
\mathcal{L}_{obj} = \frac{1}{M}\sum_{j=1}^{M}\frac{1}{|\mathcal{H}j|}\sum_{l \in \mathcal{H}j} \frac{1}{|\mathcal{O}|}
\sum_{k \in \mathcal{O}\setminus \mathcal{H}j} \\
\max\{0,\eta_{obj} + f_o^T(R_j)g(k) - f_o^T(R_j)g(l)\}
\end{multline}

\noindent
\textbf{Attribute Loss:} The attribute loss is a multi-label classification loss with two differences from object classification. Attribute labels are even less likely to be mutually exclusive than object labels. As such, we predict each attribute with independent cross entropy losses. We also weigh the samples based on fraction of positive labels in the batch to balance the positive and negative labels in the dataset. For a batch with M samples $\{(R_j,\mathcal{T}_j)\}_{j=1}^{M}$ where $\mathcal{T}_j$ is the set of attributes annotated for region $R_j$, the attribute loss is:
\vspace{-4mm}
\begin{multline}
\mathcal{L}_{atr} = \frac{1}{M}
\sum_{j=1}^{M} 
\sum_{t \in \mathcal{T}} \\
\mathbbm{1}\left[t \in \mathcal{T}_j\right](1-\Gamma(t))\log\left[\sigma(f_a^T(R_j)g(t))\right] + \\
\mathbbm{1}\left[t \notin \mathcal{T}_j\right]\Gamma(t)\log\left[1-\sigma(f_a^T(R_j)g(t))\right]
\end{multline}    
where $\sigma$ is a sigmoid activation function and $\Gamma(t)$ is the fraction of positive samples for attribute $t$ in the batch.

\begin{figure*}[ht]
\begin{center}
\includegraphics[width=\linewidth]{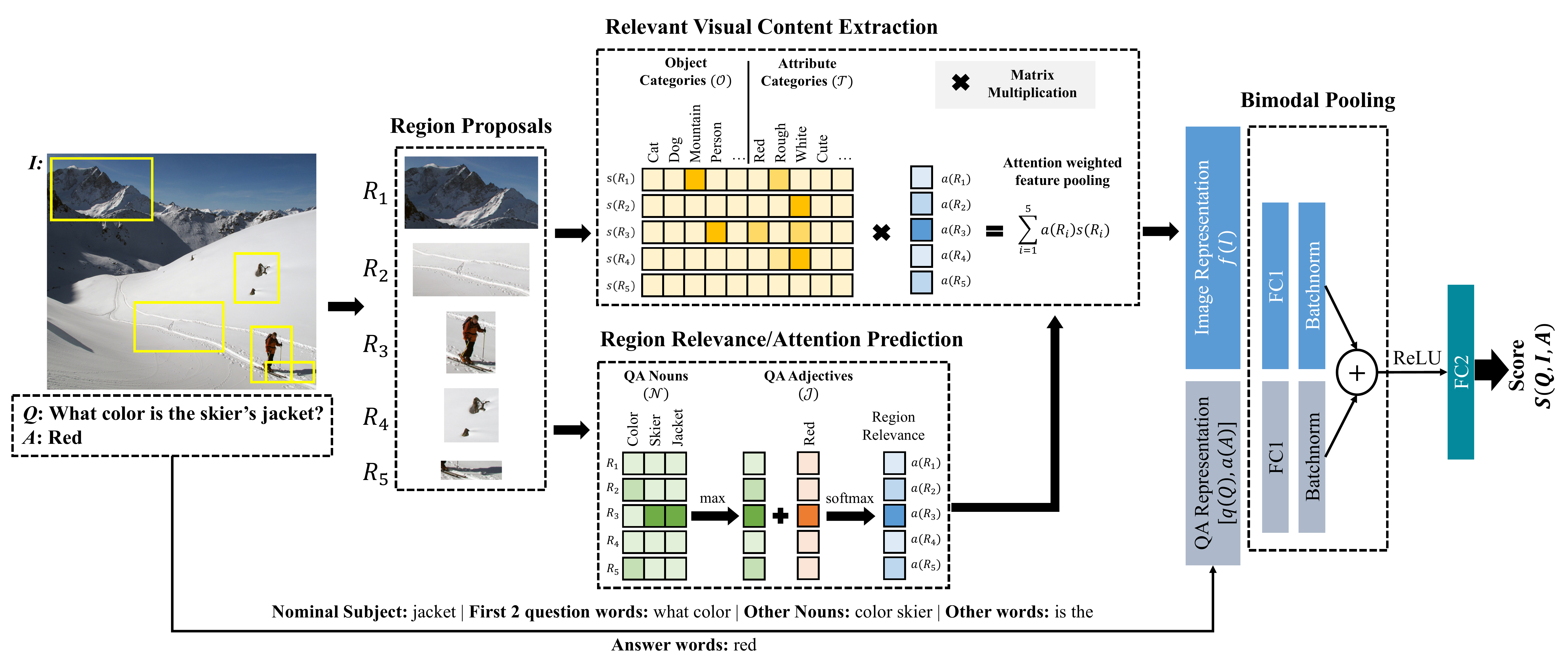}
\vspace{-1.1cm}
\end{center}
  \caption{\textbf{Inference in our VQA model:} The image is first broken down into Edge Box region proposals\cite{zitnick2014edge}. Each region $R$ is represented by visual category scores $s(R) = [s_o(R), s_a(R)]$ obtained using the visual recognition model. Using the SVLR module, the regions are also assigned an attention score using the inner products of region features with representations of nouns and adjectives in the question and answer. The region features are then pooled using the relevance scores as weights to construct the \textit{attended} image representation. Finally, the image and question/answer representations are combined and passed through a neural network to produce a score for the input question-image-answer triplet.}
  \vspace{-5mm}
\label{fig:system}
\end{figure*}
\vspace{-1mm}
\subsection{Visual Question Answering using SVLR} \label{sec:vqa}

Our VQA model is illustrated in Fig.~\ref{fig:system}.
%\subsection{Inference} \label{sec:vqa_inference}
The input to our VQA model is an image, a question, and a candidate answer. Regions are extracted from the image using Edge Boxes~\cite{zitnick2014edge}.  The same SVLR module used by VR (Sec.~\ref{sec:vr}) is explicitly applied to VQA for attention and answer scoring.  Our system assigns attention scores to each region according to how well it matches words in the question/answer, then scores each answer based on the question, answer, and attention-weighted scores for all objects ($\mathcal{O}$) and attributes ($\mathcal{T}$).

\boldhead{Attention Scoring}\label{sec:relevance} Unlike other attention models ~\cite{yang2015stacked,lu2016hierarchical} that are free to learn any correlation between regions and question/answers, our attention model encodes an explicit notion of vision-language grounding.  Let $\mathcal{R}$ be the set of region proposals extracted from the image, and $\mathcal{N}$ and $\mathcal{J}$ denote the set of nouns and adjectives in the $(Q,A)$ pair. Each region $R\in \mathcal{R}(I)$ is assigned an attention score $a(R)$ as follows:\\
\vspace{-5mm}
\begin{align}
a'(R) &= \max_{n \in \mathcal{N}} f_o^T(R)g(n) + \max_{j \in \mathcal{J}} f_a^T(R)g(j) \label{eqn:relevance_unnormalized}\\
a(R)&= \frac{\exp(a'(R))}{\sum_{R' \in \mathcal{R}(I)}\exp(a'(R'))}
\label{eqn:relevance}
\end{align}

Thus, a region's attention score is the sum of maximum adjective and noun scores for words mentioned in the question or answer (which need not be in sets $\mathcal{O}$  and $\mathcal{T}$).   

%The image representation is computed by averaging these VR scores using the attention scores as weights. This image representation is combined with the QA representation using bimodal pooling layers, and the resulting encoding of the image-question-answer triplet is scored using a fully connected layer. The answer with the highest score is chosen. \\

\boldhead{Image Representation} To score an answer, the content of region $R$ is encoded using the VR scores for all objects and attributes in $\mathcal{O}$ and $\mathcal{T}$, as presence of unmentioned objects or attributes may help answer the question. The image representation is an attention-weighted average of these scores across all regions: 
%To capture the visual content of a region $R$, we concatenate the object and attribute scores for visual categories $\mathcal{O}$ and $\mathcal{T}$ into vectors $s_o(R) \in \mathbb{R}^{1000}$ and $s_a(R) \in \mathbb{R}^{1000}$. The final 2000 dimensional, $(Q,A)$ specific, attended image representation is constructed by averaging features $s_o$ and $s_a$ across all regions using relevance scores as weights.
\begin{equation}
f(I) = \sum_{R\in\mathcal{R}(I)}a(R)
\begin{bmatrix}
s_o(R) \\
s_a(R) \\
\end{bmatrix}
\label{eqn:weightedvisualfeats}
\end{equation}
where $I$ is the image, $s_o(R)$ are the scores for 1000 objects in $\mathcal{O}$ for each image region $R$, $s_a(R)$ are the scores for 1000 attributes in $\mathcal{T}$, and $a(R)$ is the attention score.

\boldhead{Question/Answer Representation}  To construct representations $q(Q)$ and $a(A)$ for the question and answer, we follow Shih et al.~\cite{shih2016look}, dividing question words into 4 bins, averaging word representations in each bin, and concatenating the bin representations resulting in a 1200 ($=300\times4$) dimensional vector $q(Q)$. The answer representation $a(A)\in\mathbb{R}^{300}$ is obtained by averaging the word representations of all answer words. The word representations used here are produced by the SVLR module.

\boldhead{Answer Scoring} We combine the image and Q/A representations to jointly score the $(Q,I,A)$ triplet. %Since some answers like $\{0,1,2,3,yes,no\}$ may not be well represented using vector representations, we experiment with appending binary features for these answers in $a(A)$.

To ensure equal contribution of language and visual features, we apply batch normalization~\cite{batchnorm} on linear transformations of these features before adding them together to get a bimodal representation $\beta(Q,I,A)\in\mathbb{R}^{2500}$:
\begin{multline}\label{eq:bimodal_pool}
\beta(Q,I,A) = \;\mathcal{B}_1(W_1f(I)) 
+ \;\mathcal{B}_2\left(W_2 
\begin{bmatrix}
q(Q) \\
a(A) \\
\end{bmatrix}
\right)
\end{multline}
Here, $\mathcal{B}_1,\mathcal{B}_2$ denote batch normalization and $W_1\in\mathbb{R}^{2500\times2000}$ and $W_2\in\mathbb{R}^{2500\times1500}$ define the linear transformations.
\noindent
The bimodal representation is: 
\begin{equation}
\mathcal{S}(Q,I,A) = W_3 \; \text{ReLU}(\beta(Q,I,A))
\end{equation}
with  ${W_3\in\mathbb{R}^{1\times2500}}$.

\boldhead{Training}\label{sec:vqa_learn}
We use the VQA dataset~\cite{antol2015vqa} for training parameters of our VQA model: $W_1, W_2, W_3$, and scales and offsets of batch normalization layers. In addition, the VQA loss backpropagates into $f_o, f_a$, and $g$ which are part of the SVLR module. Each sample in the dataset consists of a question $Q$ about an image $I$ with list of answer options including a positive answer $A^{+}$ and $N$ negative answers $\{A^{-}(i) | i=1,\cdots, N\}$. 

The VQA loss encourages the correct answer $A^{+}$ to be scored higher than all incorrect answer options $\{A^{-}(i) | i=1,\cdots, N\}$ by a margin $\eta_{ans}$. Given batch samples $\{(Q_j,I_j,A_j)\}_{j=1}^{P}$, the loss is written as 
\vspace{-2mm}
\begin{multline}
\mathcal{L}_{ans} = \frac{1}{NP}\sum_{j=1}^{P}
\sum_{i=1}^{N} \max\{0,\\\;\eta_{ans} + \mathcal{S}(Q_j,I_j,A_j^{-}(i)) - \mathcal{S}(Q_j,I_j,A_j^{+})\}
\end{multline}

\begin{figure*}[t]
\begin{center}
\includegraphics[width=0.95\linewidth]{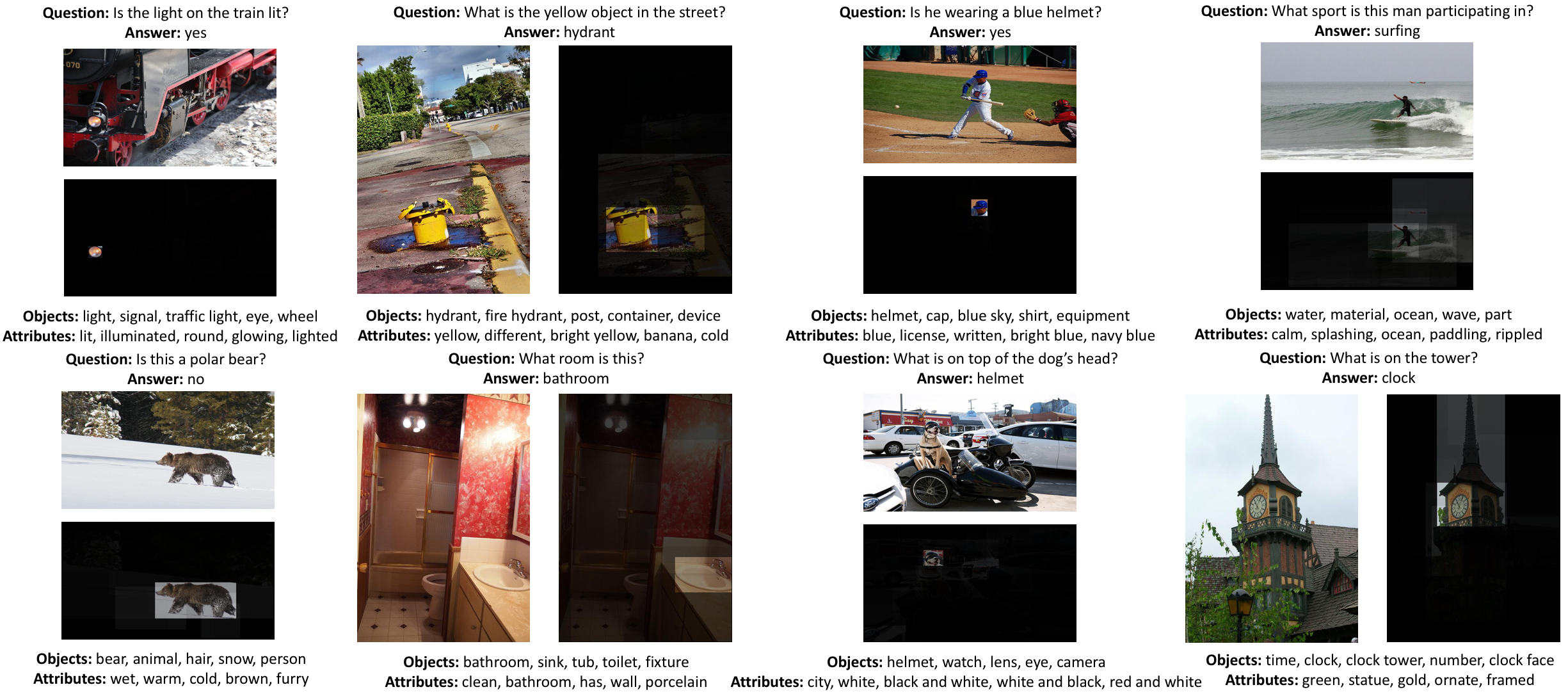}
\vspace{-0.7cm}
\end{center}
  \caption{\textbf{Interpretable inference in VQA:} Our model produces interpretable intermediate computation for region relevance and object/attribute predictions for the most relevant regions. Our region relevance explicitly grounds nouns and adjectives from the Q/A input in the image. We also show object and attribute predictions for the most relevant region identified for a few correctly answered questions. The relevance masks are generated from relevance scores projected back to their source pixels locations.}
  \vspace{-5mm}
\label{fig:rel_qual}
\end{figure*}

\subsection{Zero-Shot VQA} 

%In the introduction, we claim the ability to perform a new vision-language task without retraining as evidence that our SVLR transfers knowledge across tasks without the need to re-learn a feature to task mapping.  
The representations produced by SVLR module should be directly usable in related vision-language tasks without any additional learning. To demonstrate this \textit{zero-shot cross-task transfer}, we train the SVLR module using Genome VR data only and apply to VQA. Since bimodal pooling and scoring layers cannot be learned without VQA data, we use a proxy scoring function constructed using region-word scores only.  For each region, we compute $p_q(R)$ as the sum of its scores for the maximally aligned question nouns and question adjectives (Eq.~\ref{eqn:relevance_unnormalized} with only question words). A score $p_a(R)$ is similarly computed using answer nouns and adjectives. The final score for the answer is defined by
\begin{equation}
    S(Q,I,A)=\sum_{R\in\mathcal{R}}a(R)\min(p_q(R),p_a(R))
\end{equation}
where $a$ is the attention score computed using Eq.~\ref{eqn:relevance}. Therefore, the highest score is given to QA pairs where question as well as answer nouns and adjectives can be localized in the image. Note that the since the model is not trained on even a \textit{single question} from VQA, the zero-shot VQA task also shows that our model does use the image to answer questions instead of solely relying on the language prior which is a common concern with most VQA models \cite{agrawal2016analyzing,goyal2016arxiv}. 

\section{Implementation and Training Details}
We use 100 region proposals resized to $224 \times 224$ for all experiments. Resnet-50 was used for image feature extraction in all experiments except those in Tab.~\ref{tab:state_art} which used Resnet-152. The nouns and adjectives are extracted from the $(Q,A)$  and lemmatized using the part-of-speech tagger and WordNet lemmatizer in NLTK \cite{bird2009book}. We use the Stanford Dependency Parser \cite{de2006lrec} to parse the question into bins as detailed in~\cite{shih2016look}.  All models are implemented and trained using TensorFlow \cite{tensorflow2015software}. We train the model jointly for the recognition and VQA tasks by minimizing the following loss function using Adam~\cite{adamoptimizer}:
\begin{equation}
\mathcal{L} = \alpha_{ans}\mathcal{L}_{ans} + \alpha_{obj}\mathcal{L}_{obj} + \alpha_{atr}\mathcal{L}_{atr}
\end{equation}
We observe that values of $\alpha_{obj}$ and $\alpha_{atr}$ relative to $\alpha_{ans}$ can be used to trade-off performance between visual recognition and VQA tasks. For experiments that analyze the effect of transfer from VR to VQA (Sec.~\ref{sec:vr2vqa}), we set ${\alpha_{ans} = 1, \alpha_{obj} = 0.1}$, and ${\alpha_{atr}=0.1}$. For VQA only and Genome only baselines, we set the corresponding $\alpha$ to 1 and others to 0.  For experiments dealing with transfer in the other direction (Sec.~\ref{sec:vqa2vr}), we set ${\alpha_{ans} = 0.1, \alpha_{obj} = 1}$, and ${\alpha_{atr}=1}$.  The margins used for object and answer losses are $\eta_{ans}=\eta_{obj}=1$. The object and attribute losses are computed for the same set of Visual Genome regions with a batch size of $M=200$. The answer loss is computed for a batch size of $P=50$ questions sampled from VQA. We use an exponentially decaying learning rate schedule with an initial learning rate of $10^{-3}$ and decay rate of 0.5 every 24000 iterations. Weight decay is used on all trainable variables with a coefficient of $10^{-5}$. All the variables are Xavier initialized \cite{glorot2010aistats}.

\section{Experiments}
\label{sec:experiments}
Our experiments investigate the extent to which using SVLR as a core representation improves transfer in multitask learning.  We first analyze how including the VR task improves VQA (Sec.~\ref{sec:vqa_eval}, Tab.~\ref{Tbl:abltionperf}).  We find that using SVLR doubles the improvement compared to standard multitask learning, and demonstrate performance well above chance in a zero-shot setup (trained only on VR, applied to VQA).  We then analyze improvement to VR due to training with (weakly supervised) VQA (Sec.~\ref{sec:vqa2vr}, Fig.~\ref{fig:obj_change}).  We find moderate overall improvements (1.2$\%$), with the largest improvements for classes that have few VR training examples.  We also quantitatively evaluate how well our attention maps correlate with that of humans using data provided by~\cite{das2016human} in Table~\ref{tbl:vqahat}. We include results of our VQA system trained with ResNet-152 architecture on val, test-dev, test-std, along with state-of-the-art (Tab.~\ref{tab:state_art}). 

\begin{table*}
    \centering 
\setlength\tabcolsep{4 pt}

    \resizebox{\textwidth}{!}{
    \begin{tabular}{|c||c|c|c|c|c||c|c|c|c|c|c|c|c|c|c|c||c|c|c||c|}
    \hline
     %\textbf{\parbox{4cm}{Accuracies on Real-MCQ-VQA Validation Set}}&
     \textbf{Accuracies on Real-MCQ-VQA Validation Set}&
     \rot{\textbf{what color}}& 
     \rot{\textbf{\parbox{3cm}{what is the\\(wo)man/person}}}& 
     \rot{\textbf{what is in/on}}& 
     \rot{\textbf{\parbox{3cm}{what kind/\\type/animal}}}&
     \rot{\textbf{what room/sport}}&
     \rot{\parbox{3cm}{can/could/\\does/do/has}} & 
     \rot{\parbox{3cm}{what does/\\number/name}}& 
     \rot{what brand}&
     \rot{which/who}& 
     \rot{what is/are}& 
     \rot{why/how}&
     \rot{how many}&
     \rot{what time}& 
     \rot{where}&
     \rot{is/are/was}&
     \rot{none of the above}&
     \rot{other} &
     \rot{number}&
     \rot{yes/no}&
     \rot{overall accuracy}\\ \hline
     VQA Only & 53.5 & 70.5 & 53.6 & 56.8 & 89.8 & 81.8 & 41.9 & 45.9 & 49.0 & 58.3 & \textbf{33.8} & 38.4 & \textbf{53.9} & 45.8 & 80.2 & 56.0 & 54.5 & 39.2 & 82.1 & 62.9 \\
     \parbox{5cm}{\centering Joint Multitask} & 59.4 & 71.8& 54.6 & 58.3 & 91.0 & 81.9 & \textbf{43.8} & 46.4 & 50.8 & 59.2 & 32.3 & \textbf{39.4} &  \textbf{53.9} & 47.0 & 80.4 & 57.1 & 56.7 & \textbf{39.8} & 82.2 & 64.1 \\
     \parbox{5cm}{\centering Joint SVLR} & \textbf{62.1} & \textbf{74.1} & \textbf{57.9} & \textbf{60.0} & \textbf{91.1} & \textbf{82.8} & 41.6 & \textbf{52.9} & \textbf{52.0} & \textbf{61.1} & 33.6 & 39.0 & 51.3 & \textbf{48.6} & \textbf{81.4} & \textbf{58.5} &  \textbf{58.8} & 38.8 & \textbf{83.0} & \textbf{65.3} \\ \hline
     \parbox{5cm}{\centering Zero-Shot VQA} & 18.8 & 21.0 & 27.4 & 31.4 & 22.0 & 17.1 & 13.9 & 11.6 & 20.6 & 22.9 & 12.7 & 0.7 & 7.2 & 26.1 & 13.5 & 19.2 & 22.4 & 1.2 & 13.3 & 16.4 \\ \hline
     %\parbox{5cm}{\centering Zero-Shot VQA} & 36.5 & 28.7 & 37.6 & 33.6 & 55.6 & 6.1 & 13.0 & 7.6 & 34.2 & 39.6 & 16.2 & 3.8 & 14.1 & 42.5 & 9.5 & 27.0 & 34.4 & 5.0 & 8.1 & 20.8 \\ \hline
    \end{tabular}}
    \vspace{-3mm}
    \caption{\textbf{Inductive transfer from VR to VQA through SVLR in joint training and zero-shot settings:} We evaluate the performance of our model with SVLR module trained jointly with VR and VQA supervision (provided by Genome and VQA datasets respectively) on the VQA task. We compare this \textit{jointly-trained} model to a model trained on \textit{only} VQA data. We also compare to a traditional multitask learning setup that is jointly trained on VQA and VR (i.e. uses same amount of data as Joint SVLR) and shares visual features but \textit{does not} use the object and attribute word embeddings for recognition. While multitask learning outperforms VQA-only model, using the SVLR module doubles the improvement. Our model is most suited for the question types in bold that require visual recognition without specialized skills like counting or reading. Formulation of VR and \textit{attention} in VQA in terms of inner products between word and region representations enables Zero-Shot VQA. In this setting we train on Genome VR data and apply to VQA val (Sec~\ref{sec:vqa_eval}).}
    \vspace{-3mm}
    \label{Tbl:abltionperf}
\end{table*}

\subsection{Datasets}
Our model is trained on two separate datasets: one for VQA supervision, one for visual recognition (attributes and object classification). We use the image-question-answer annotation triplets from Antol et al. \cite{antol2015vqa} and bounding box annotations for object and attribute categories from Visual Genome~\cite{krishna2016visual}. The train-val-test splits for the datasets are as follows. 

\boldhead{VQA} We split the \textit{train} set into \textit{train-subset} and \textit{train-held-out} and use the latter for model selection. The \textit{train-subset} consists of 236,277 $(Q,I,A)$ samples whereas \textit{train-held-out} contains 12,072 samples. The \textit{val} and \textit{test} set contain 121,512 and 244,302 samples respectively. There are exactly 3 questions per image. We use VQA val for evaluating on specific question types.

\boldhead{Visual Genome} We use only images from Visual Genome not in VQA (overlaps identified using md5 hashes). The selected images were divided into \textit{train}-\textit{val}-\textit{test} using an 85-5-10 split, yielding 1,565,280, 90,212 and 181,141 annotated regions in each. We use \textit{val} for selecting the model for evaluating recognition performance. 

\subsection{Inductive Transfer from VR to VQA}\label{sec:vqa_eval}\label{sec:vr2vqa}
In Table~\ref{Tbl:abltionperf}, we analyze the role of SVLR module for inductive transfer in both joint training and zero-shot settings.

\boldhead{Joint Training} During joint training, the VR models and VQA model are simultaneously trained using object and attribute annotations from Genome, and Q/A annotations from the VQA dataset. The common approach to joint training is to use a common network for extracting image features (e.g. class logits from ResNet), which feeds into the task-specific networks as input. We refer to this approach in Table~\ref{Tbl:abltionperf} as \textit{Joint Multitask}. This baseline is implemented by replacing $g(y)$ (see Fig.~\ref{fig:features}),  with a fixed set of vectors $h_y$ for each of the predetermined 1000 object and 1000 attribute categories in the VR models. The embedding $g(y)$ is still in the VQA model, but is no longer shared across tasks. Our proposed \textit{Joint SVLR} outperforms VQA-only by $2.4\%$, doubling the $1.2\%$ improvement achieved by \textit{Joint Multitask}.  Our formulation of VR and VQA tasks in terms of shared word-region representations more effectively transfers recognition knowledge from VR than shared features.  The gain is often larger on questions that involve recognition (in bold in Table~\ref{Tbl:abltionperf}).  For example, \textit{what color} questions improve by $8.6\%$ due to SVLR.

Surprisingly, pre-training the visual classifiers on Genome prior to joint training performs worse than the model trained jointly from scratch: 63.7\% versus 65.3\%.

\boldhead{Zero-Shot VQA} We evaluate Zero-shot VQA to further highlight transfer from VR to VQA. We train on only Genome VR annotations but test on VQA val. The model has not seen any Q/A training data, but achieves an overall accuracy of $16.4\%$ where random guessing yields $5.6\%$ (18 choices).  Our zero-shot system does not exploit language priors, which alone can score as high as 54.0$\%$~\cite{shih2016look}. This shows that some knowledge can be directly applied to related tasks using SVLR without additional training.
%Our result demonstrates that given our formulation, knowlege from the VR task can directly contribute to the performance of VQA without the need to learn a feature to task mapping.\\

\begin{figure}
\vspace{-2mm}
\includegraphics[width=0.95\columnwidth]{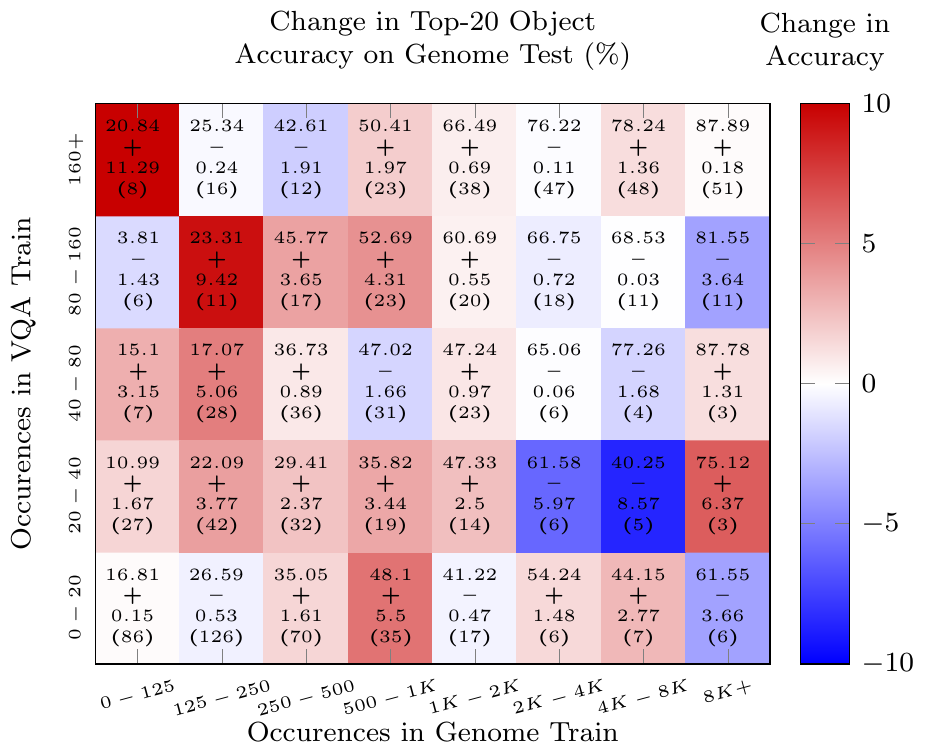}
\vspace{-4mm}
\caption{\textbf{Transfer from VQA to Object Recognition:} Each cell's color reflects the mean change in accuracy for classes within the corresponding frequency ranges of both datasets' training split. Most gains are in nouns rare in Genome but common in VQA (top left), suggesting that the weak supervision provided by training VQA attention augments recognition performance via the SVLR. The numbers in each cell show the Genome-only mean accuracy +/- the change due to SVLR multitask training, followed by the number of classes in the cell in parentheses.}
\vspace{-2mm}
\label{fig:obj_change}
\end{figure}

%As shown in Table~\ref{tab:state_art}, our model achieves competitive results on VQA. Our model offers significantly improved localization performance as implied by \textit{val} set performance on \textit{color} questions. This is because questions about color require precise localization of the relevant area to make an accurate classification.\\ \todo{TODO: update with 152 numbers}

\begin{figure*}[ht]
\begin{center}
\includegraphics[width=0.95\linewidth]{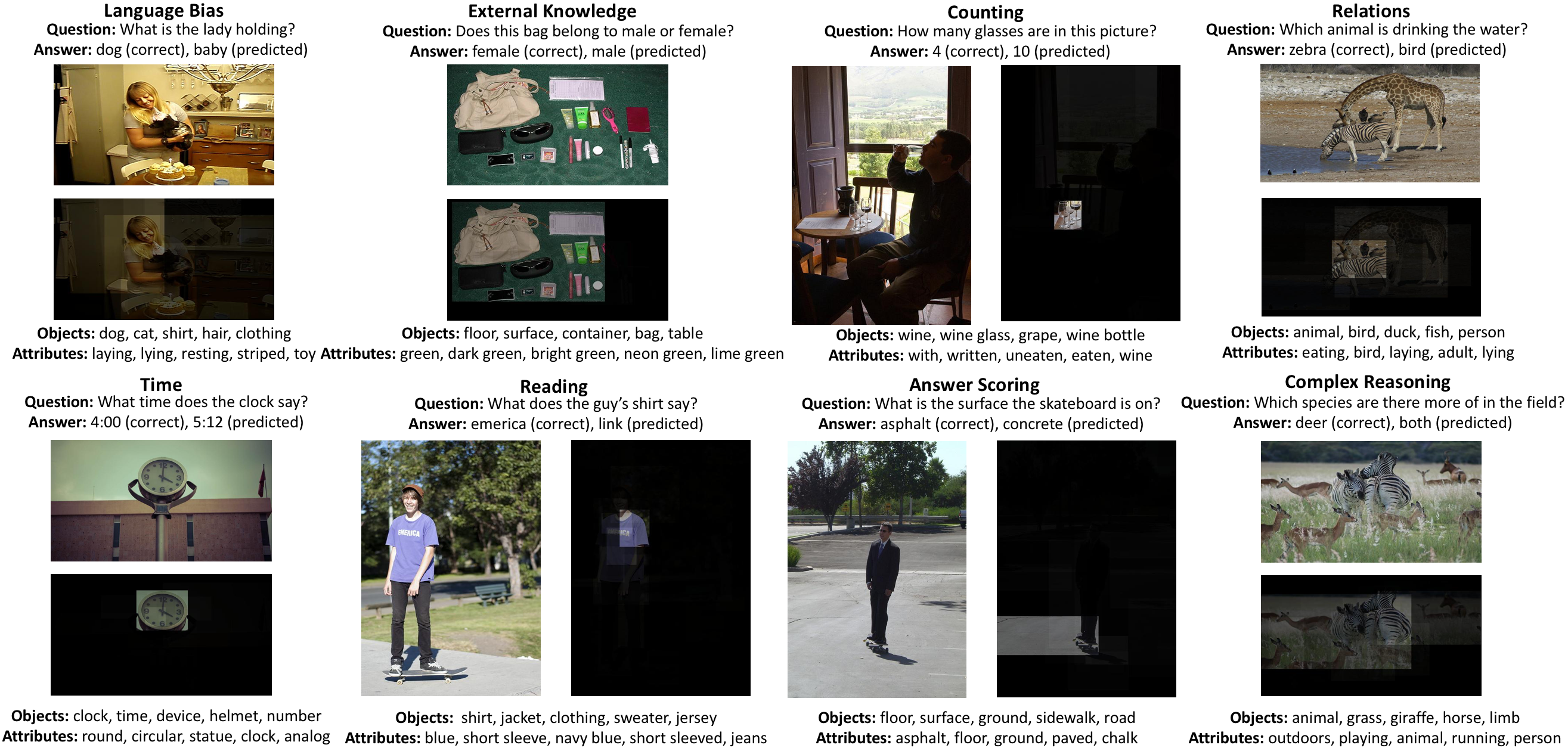}
\vspace{-0.7cm}
\end{center}
  \caption{\textbf{Failure modes:} Our model cannot count or read, though it will still identify the relevant regions. It is blind to relations and thus fails to recognize that \textit{birds}, while present, are not \textit{drinking water}. The model may give a low score to the correct answer despite accurate visual recognition. For instance, the model observes \textit{asphalt} but predicts \textit{concrete}, likely due to language bias. A clear example of an error due to language bias is in the top-left image as it believes the lady is holding a \textit{baby} rather than a \textit{dog}, even though visual recognition confirms evidence for dog. Finally, our model fails to answer questions that require complex reasoning comparing multiple regions.}
  \vspace{-2mm}
\label{fig:fail_modes}
\end{figure*}

\subsection{Inductive Transfer from VQA to VR}\label{sec:vqa2vr}
We compare the performance of our SVLR based model trained jointly on VQA and VR data with a model trained only on Genome data to analyze transfer from VQA to VR. Genome \textit{test} is used for evaluation. We observe an increase in the overall object recognition accuracy from 43.3\% to 44.5\%, whereas average attribute accuracy remained unchanged at 36.9\%. In Fig.~\ref{fig:obj_change}, we show that nouns that are rare in Genome (left columns) but have 20 or more examples in VQA (upper rows) benefit the most from weak supervision provided by VQA.  On average, we measure improvement from 21$\%$ to 32$\%$ for the 8 classes that have fewer than 125 examples in Genome train but occur more than 160 times in VQA questions. We conducted the same analysis on Genome attributes, but did not observe any notable pattern, possibly due to the inherent difficult in evaluating the multi-label attribute classification problem (the absence of attributes is not annotated in Genome).

%For object recognition, the accuracy for object classes which are neither too rare nor too frequent in the Genome training set improve while performance on most others is preserved. We do not expect to see improvement for classes which are already well represented in the Genome training data, nor for classes too rare for both datasets (bottom left cell). This effect is observed to a lesser extent in case of attributes, but we believe it is due to incomplete attribute annotations in the Genome dataset that prevent accurate multi-label evaluation. But we include attribute recognition as one of the tasks during joint training to enable VQA model to answer attribute based questions.   

\subsection{Interpretable Inference for VQA}
As shown in Fig. \ref{fig:rel_qual}, our VQA model produces interpretable intermediate outputs such as region relevance and visual category predictions, similar to \cite{tommasi2016bmvc}. The answer choice is explained by the object and attribute predictions associated with the most relevant regions. Because relevance is posed as the explicit localization of words in the question and answer, we can qualitatively evaluate the relevance prediction by verifying that the predicted regions match said words. This also provides greater insight into the failure modes as shown in Fig. \ref{fig:fail_modes}. 

\begin{figure}                                  
  \centering                                                                        
  \includegraphics[width=\columnwidth]{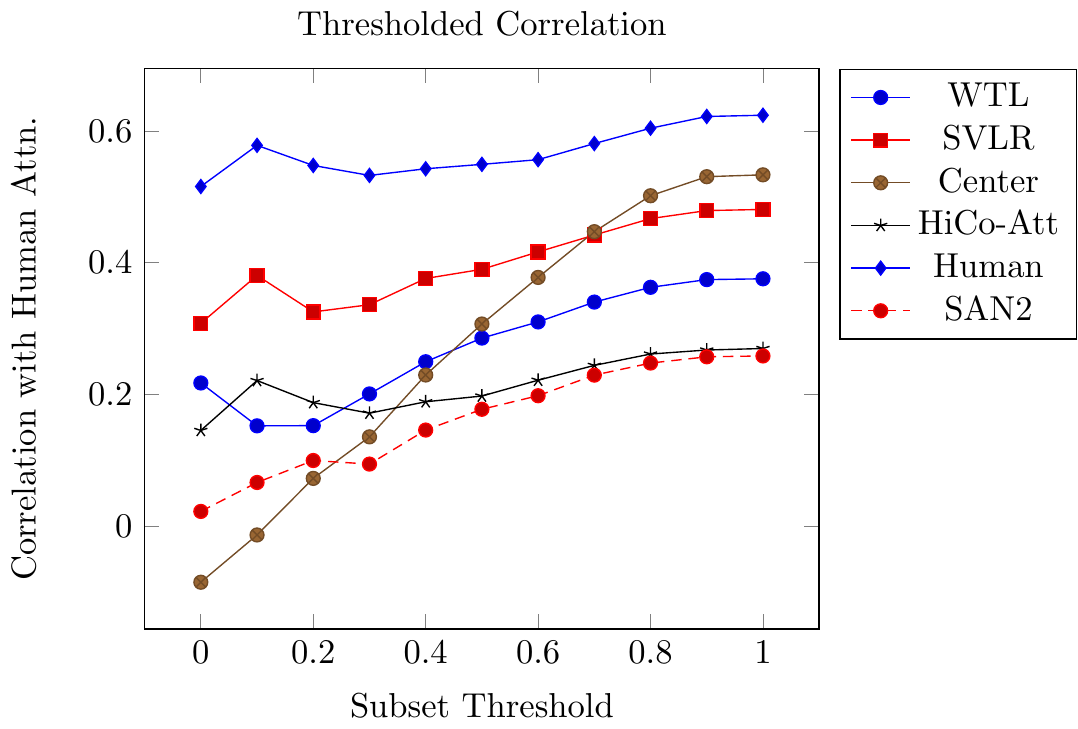}  
  \caption{\textbf{Mean Spearman rank-correlation between model
      predicted and human attention at various thresholds.} Each
      threshold point defines a subset of the dataset for
      which the human attention correlation with the synthetic center
      heatmap is below that threshold value. For example: the first sample
      point of each curve is the mean correlation of each model with
      human attention, measured on a subset in which the human
      attention's correlation with the center heatmap is
      less than or equal to 0. As can
    be seen, the attention maps produced by the proposed SVLR model correlate with human attention significantly more than other models. As the threshold
    approaches 1, the synthetic center heatmap baseline outperforms
    all proposed models, confirming that the majority of the questions
    are about something in the center of the image. Note that due to slight differences in implementations, the subsets at $\le 0$ differ slightly from those used in \cite{das2016human}}
  \label{fig:vqahat_comparison}                                                                                                     
\end{figure}

We also quantitatively evaluate our attention using collected human attention maps from Das et al.~\cite{das2016human} in Table~\ref{tbl:vqahat} and in Figure~\ref{fig:vqahat_comparison}. Table~\ref{tbl:vqahat} includes the correlation scores between the attention maps from various VQA attention models and human attention on a subset of question-answer pairs on the validation set. Our proposed SVLR model significantly outperforms other models we compare with. However, we note that a strong center-focused heatmap baseline still outperforms all models, signifying that the main topic of a question is very often located in the center of the image. As such, we also evaluate correlations on multiple subsets of the human attention maps is Figure~\ref{fig:vqahat_comparison}, thresholding them based on correlation with the center heatmap. We note that learned attention models appear to have better correlation with human attention at lower thresholds where the human attention correlates poorly with the center-focused heatmap -- a result also demonstrated in ~\cite{das2016human}.

%We also quantitatively evaluate our attention using collected human attentions from Das et al.~\cite{das2016human} in Table~\ref{tbl:vqahat}. The comparison with WTL~\cite{shih2016look} is most informative, as it generates its attention map more similarly to the SVLR than HiCo~\cite{liu2016attention} and SAN~\cite{yang2015stacked}, specifically in the use of weighted Edge Boxes.
%
\begin{table}
\setlength{\tabcolsep}{3 pt}
\small
    \centering
    \footnotesize
    \begin{tabular}{|c|c|c|c|c||c|c|}
        \hline
         & HiCo\cite{lu2016hierarchical} & SAN2\cite{yang2015stacked}& WTL\cite{shih2016look} & SVLR & Center & Human \\
         \hline
         Corr.& 0.27 & 0.26 & 0.38 & \textbf{0.48} & 0.53 &0.62\\
         \hline
    \end{tabular}
    \vspace{-1em}
    \caption{\textbf{Human Attention Comparison:} We compare our attention maps with human attentions collected by Das et al.~\cite{das2016human}. Comparison was done by resizing attention maps to 14$\times$14 and computing the Spearman rank correlation as in ~\cite{das2016human}. We include a strong baseline using a synthetic center-focused heatmap (also used by ~\cite{das2016human}) under the Center column. The Human column represents the inter-annotator agreement. Scores for HiCo and SAN2 were recomputed using released data from~\cite{das2016human}, and differ slightly from originally reported. Our model leads to significantly higher correlation with human annotations than existing models.}
    \label{tbl:vqahat}
    \vspace{-2em}
\end{table}

\subsection{Learned Word Representations}
In Table~\ref{Tbl:embeddingexamples}, we compare the word representations of the SVLR model to that of Word2Vec~\cite{word2vec} by showing several nearest neighbors from both embeddings. We observe a shift from non-visual neighborhoods and meanings (monitor, control) to visual ones (monitor, keyboard). Neighbors were computed using cosine distance after mean centering. 

\begin{table}[]
\scriptsize
\centering
\setlength{\tabcolsep}{2 pt}
\resizebox{\columnwidth}{!}{
\begin{tabular}{|c|c|c|c|c|c|c|}
\hline 
& WTL\cite{shih2016look}  & FDA\cite{ilievski2016focused}  & MLP\cite{mallya2016simplevqa,jabri2016revisiting}  &  MCB\cite{fukui2016multimodal} & HiCo\cite{lu2016hierarchical} & Ours \\

\hline
%\multicolumn{2}{|c|}{Base recognition network}                                                                       & VGG         & Resnet-101 & Resnet-152  & Resnet-152    & Resnet-200 & Resnet-152          \\\hline
%\multirow{6}{*}{\rot{\parbox{4cm}{\centering val}}} & What color (9.8\%)                                                            & 54.0          & -    & 51.9  & - & - & 65.0                      \\
%                     & What is in/on (1.8\%)                                                         & 54.8          & -    & 61.7 & - & -  & 60.1                      \\
%                     & \begin{tabular}[c]{@{}c@{}}What kind/\\ type/animal (23.8\%)\end{tabular}     & 52.9        & -    & 65.8  & - & -  & 61.1                      \\
%                     & \begin{tabular}[c]{@{}c@{}}What is the \\ man/woman/person (2\%)\end{tabular} & 70.2        & -    & 78.0    & - & -  & 76.1                      \\
%                     & \begin{tabular}[c]{@{}c@{}}Can/could/does/do/has \\ (4.6\%)\end{tabular}      & 75.7       & -    & 51.5 & -  & -  & 82.6                  \\
val                                                                      & 58.9         & -    & 63.6  & -  & -  & 66.2 \\  \hline
test-dev                                                                       & 62.4           & 64.0 & 65.9  & 69.9 & 65.8 & 64.8                      \\
test-std                                                                       & 63.5          & 64.2 & -  & -    & 66.1 & 64.8          \\ \hline
Trained on                                                                        & \textit{train+val}          & \textit{-} & \textit{train}  & \textit{train+val}    & \textit{train+val} & \textit{train}  \\\hline 
\end{tabular}}
\vspace{-2mm}
\caption{\textbf{External Comparisons on VQA:} We include external comparisons, but note that internal comparisons are more controlled and informative.  The MLP results use the implementation from \cite{mallya2016simplevqa}. For test accuracy, it is unclear whether FDA uses \textit{val} to train. The original MLP implementation \cite{jabri2016revisiting} using Resnet-101 yields 64.9 and 65.2 on \textit{test-dev} and \textit{test-std} respectively. MCB reports only \textit{test-dev} accuracy for the directly comparable model (final without ensemble). Note that the overall performance of our model is slightly worse than MLP and MCB because only about $10\%$ of the VQA dataset benefits from visual attention. Our model achieves $62.1\%$ on color questions using attention, outperforming WTL's $54\%$ and MLP's $51.9\%$.}
\label{tab:state_art}
\end{table}

\vspace{-3mm}
\begin{table}[t]
\scriptsize
    \centering
    \setlength\tabcolsep{2pt}
    \begin{tabular}{|c|c|c|}
    \hline
    Word & Word2Vec & SVLR \\
    \hline
    %band & guitar, rock, group, piano & wristband, sock, headband, collar \\
    column & newspaper, magazine, book, letter & pillar, post, pole, tower, chimney \\
    counter & curb, stem, foil, stop, dispenser & shelf, stove, countertop, burner \\
    horn & piano, guitar, brass, pedal & tail, harness, tag, paw \\
    meat & chicken, lamb, food, uncooked & rice, scrambled, piled, slice \\
    monitor & watch, control, checked, alert & keyboard, computer, portable \\
    \hline
    \end{tabular}
    \vspace{-2mm}
    \caption{\textbf{Word Representations from SVLR vs Word2Vec:} We compare nearest neighbors (cosine distance) for a set of words using word2vec embeddings as well as SVLR.}
    \label{Tbl:embeddingexamples}
    \vspace{-3mm}
\end{table}
\section{Conclusion}
%We propose an effective means of transferring knowledge and representations between related vision-language tasks by formulating each tasks' model around a shared image-to-text embedding space (SVLR). By sharing knowledge across tasks in the form of alignments between visual features and text, the shared representation has the same interpretation across all tasks, allowing supervision in one task to directly benefit the other by means of improving the shared embedding space. 

Humans learn new skills by building upon existing knowledge and experiences. We attempt to apply this behavior to AI models by demonstrating cross-task learning for the class of vision-language problems using VQA and VR. To enhance inductive transfer, we propose sharing core vision and language representations across all tasks in a way that exploits the word-region alignment. We plan to extend our method to larger sets of vision-language tasks. 

%We designed an interpretable Vision-Language representation space that can be reused across multiple tasks. We show learning and application of the shared representation space through visual recognition and the VQA tasks. We also formulate the VQA task in terms of the recognition task which allows more efficient utilization of recognition supervision for VQA and weak supervision of visual recognition through VQA. As future work, we plan to extend the approach to other tasks such as activity recognition and image-caption retrieval. For VQA, we plan to model pairwise relations explicitly to answer questions involving spatial reasoning. 

\section{Acknowledgements}
This work is supported in part by NSF Awards 14-46765 and 10-53768 and ONR MURI N000014-16-1-2007.

{\small
\bibliographystyle{ieee}
\bibliography{egbib}
}

\end{document}